\documentclass[9.5pt,journal,compsoc,cspaper]{IEEEtran}

\usepackage[pdftex]{graphicx}
\usepackage{arydshln}
\usepackage{booktabs, makecell, tabularx}
\usepackage[export]{adjustbox}
\usepackage{array}
\usepackage{ragged2e}
\usepackage[caption=false,font=normalsize,labelfont=sf,textfont=sf]{subfig}
\usepackage{textcomp}
\usepackage{stfloats}
\usepackage{url}
\usepackage{multirow}
\usepackage{verbatim}
\usepackage{bm} 
\usepackage[table]{xcolor}
\usepackage{colortbl}
\usepackage{pifont}
\usepackage{arydshln}
\usepackage{graphicx}
\usepackage{cite}
\usepackage[pagebackref=true,breaklinks=true,letterpaper=true,colorlinks,bookmarks=false]{hyperref}

\definecolor{todocolor}{RGB}{0,174,247}
\definecolor{myblue}{RGB}{0,165,255}
\definecolor{myyellow}{RGB}{255,215,0}

\ifCLASSINFOpdf
  \usepackage[pdftex]{graphicx}
  \graphicspath{{../pdf/}{../jpeg/}{./figures}}
\else

\fi

\usepackage{xspace}
\usepackage{pgfplots}
\usepackage{xcolor}
\usepackage[table]{xcolor}
\usepackage{tikz}

\usepackage{amsmath,amsfonts}
\usepackage{amssymb}

\renewcommand{\baselinestretch}{.985}

\begin{document}

\title{Detecting Every Object from Events}

\author{
    Haitian Zhang, Chang Xu, Xinya Wang, Bingde Liu, Guang Hua, Lei Yu, Wen Yang
}

\markboth{Journal of \LaTeX\ Class Files,~Vol.~XX, No.~XX, April~2024}%
{Shell \MakeLowercase{\textit{et al.}}: Bare Demo of IEEEtran.cls for Computer Society Journals}

\newcommand{\ie}{\textit{i.e.}\xspace}
\newcommand{\eg}{\textit{e.g.}\xspace}
\newcommand{\etal}{\textit{et al.}\xspace}
\newcommand{\wrt}{\textit{w.r.t.}\xspace}
\newcommand{\etc}{\textit{etc.}\xspace}
\newcommand{\resp}{\textit{resp.}\xspace}

\definecolor{todocolor}{RGB}{0,174,247}

\IEEEtitleabstractindextext{%
\begin{abstract}
    \justifying
Object detection is critical in autonomous driving, and it is more practical yet challenging to localize objects of unknown categories: an endeavour known as Class-Agnostic Object Detection (CAOD). 
Existing studies on CAOD predominantly rely on ordinary cameras, but these frame-based sensors usually have high latency and limited dynamic range, leading to safety risks in real-world scenarios.
In this study, we turn to a new modality enabled by the so-called event camera, featured by its sub-millisecond latency and high dynamic range, for robust CAOD. We propose Detecting Every Object in Events (DEOE), an approach tailored for achieving high-speed, class-agnostic open-world object detection in event-based vision. Built upon the fast event-based backbone: recurrent vision transformer, we jointly consider the spatial and temporal consistencies to identify potential objects. The discovered potential objects are assimilated as soft positive samples to avoid being suppressed as background.
Moreover, we introduce a disentangled objectness head to separate the foreground-background classification and novel object discovery tasks, enhancing the model's generalization in localizing novel objects while maintaining a strong ability to filter out the background.
Extensive experiments confirm the superiority of our proposed DEOE in comparison with three strong baseline methods that integrate the state-of-the-art event-based object detector with advancements in RGB-based CAOD. Our code is available at \url{https://github.com/Hatins/DEOE}.
\end{abstract}

}

\maketitle

\IEEEpeerreviewmaketitle


\section{Introduction}
\label{sec:introduction}

\IEEEPARstart{E}{nvision} the following scenario, where a car is speeding towards an intersection in the darkness, while simultaneously, a fast-moving object suddenly emerges from a nearby junction, leading to safety hazards.
To avoid accidents, pivotal details encompassing the object's identity, its motion state, size, and distance from the vehicle are required for the autonomous driving perception system. 
However, all of these prerequisites are contingent upon the swift and precise perception of the fast-moving object, represented by bounding box predictions \cite{frame_caod_2021_RAL}. This scenario is not limited to the autonomous driving field, instead, this prototype illustrates the challenge raised by fast-moving objects across various domains, including object tracking \cite{event_track_2018_IROS}, obstacle avoidance \cite{UAV_2016_TPAMI}, and related applications.

To assist the decision system in making quick and accurate responses, the perception system should be able to localize every object captured by the vision sensor within a short time interval, regardless of the environmental conditions. 
The event camera \cite{event_survey_2020_TPAMI}, distinguished by its sub-millisecond temporal resolution and broad dynamic range, provides potential robustness for swiftly detecting every object under extreme conditions \cite{event_dataset_2020_NIPS, event_detector_2022_TIP, event_detector_2023_CVPR}, \textit{e.g.}, the aforementioned fast-moving object scenario, over exposure, and darkness environments.

\begin{figure}[!t]
    \centering
    \includegraphics[width=3.3in]{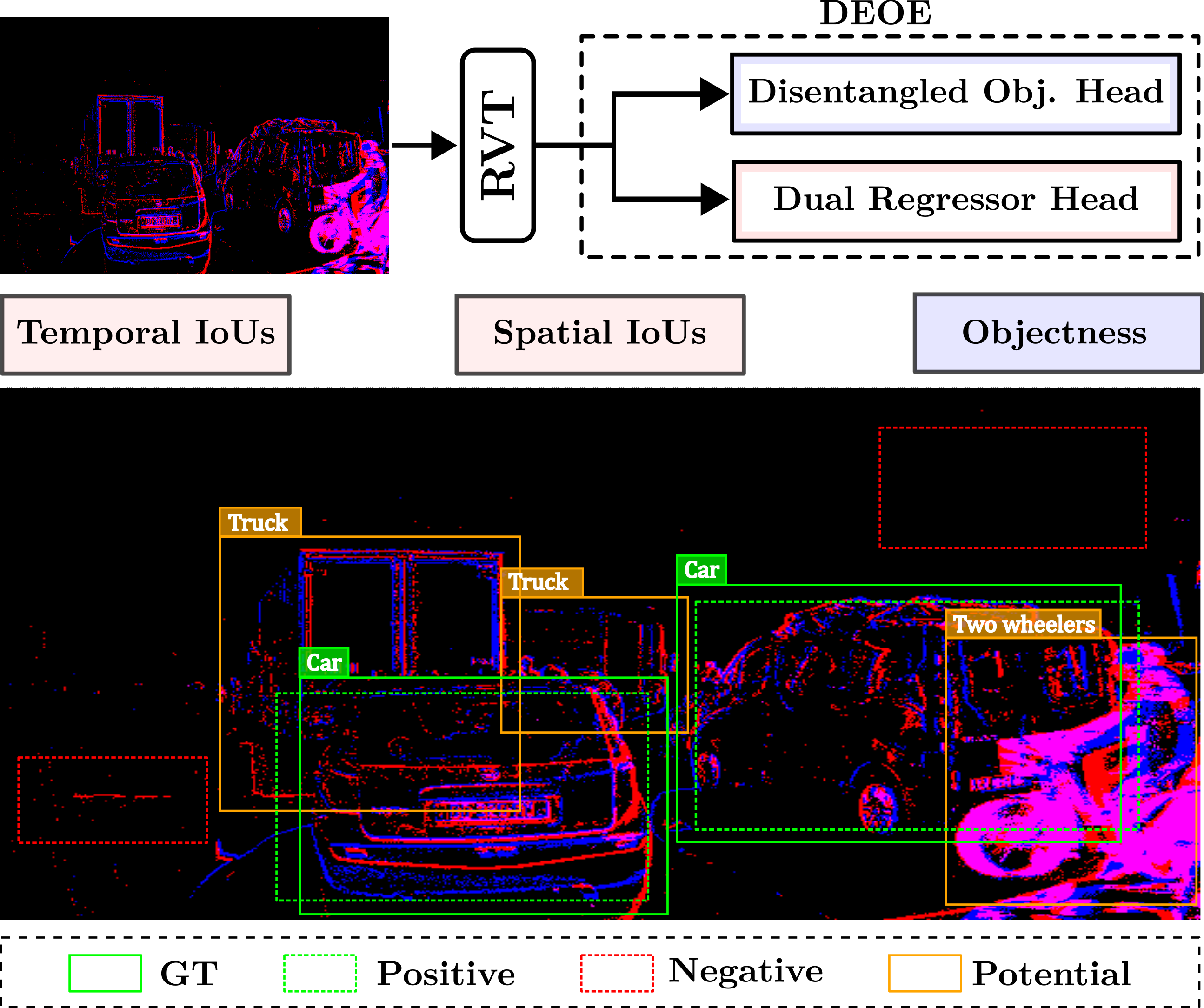}
    \caption{An overview of the DEOE (upper) and illustration of the sampling process in CAOD (lower), note that the entire model is built upon a rapid event-based backbone, recurrent vision transformer (RVT) \cite{event_detector_2023_CVPR}. Upper: DEOE consists of the \textit{Disentangled Objectness Head} and the \textit{Dual Regressor Head}. These two heads perform their tasks while simultaneously providing the metrics ``objectness", ``Spatial IoU", and ``temporal IoU", which are used to identify potential object samples. Lower: In object detection, it's common practice to assign anchors with a high IoU value with the ground truth (GT) as positive samples (green dashed boxes), while those with low IoU values are defined as negative samples (red dashed boxes). However, in the context of CAOD, anchors containing unknown objects may lack annotations and be inappropriately treated as negative samples, referred to as potential samples (orange boxes). In this example, we assume that only ``car" is annotated, whereas ``trucks" and ``two-wheelers" lack annotations.}
    \label{False_negative}
\end{figure}

Event-based object detectors have demonstrated their efficiency and robustness under a closed-set hypothesis \cite{event_survey_2020_TPAMI, event_dataset_2020_NIPS, event_detector_2023_CVPR}, but towards a more generic open-world perception, the detector should have the capability to localize not only known (annotated) objects, but also unknown (unannotated) ones, known as the Class-Agnostic Object Detection (CAOD) task, which is currently dominant by RGB-based methods\cite{frame_caod_2021_RAL,frame_caod_2022_ECCV,frame_owod_2021_CVPR,frame_owod_2022_CVPR,frame_owod_2023_CVPR}.
However, to fully harness the potential of event cameras, it is essential to leverage their continuous temporal information and high-speed characteristics, which are understudied in related CAOD research.

Towards this end, we design a method to Detect Every Object from Events (DEOE), as shown in Fig.~\ref{False_negative}. Unlike frame-based sensors, event camera captures intensity changes whenever they occur, forming continuous event streams that encode time, location, and brightness information of objects. Therefore, we aim to leverage the temporal, spatial, and objectness clues for better discovery of unknown objects in the event stream. Subsequently, we employ these discovered unknown objects to further enhance the model's discrimination capabilities.

Specifically, we first measure the foreground probability of samples via a joint consideration of temporal and spatial consistencies and enhance the positive supervision signal of unknown objects. An inherent property of objects in event streams is that they tend to appear continuously for a while \cite{event_dataset_2020_NIPS}, especially in the autonomous driving scenario. Thus, we enhance those samples that yield consistent location predictions (measured by IoU) across different event stamps during training. On the other hand, we use the IoU between dual-head predictions to measure the spatial consistency for object discovery, based on the intuition that different models will make similar predictions for semantically consistent items. 
Second, we disentangle foreground-background classification and unknown object discovery via decoupling the learning targets by different objectness heads. Working in a \textit{dividing-and-conquering} manner, one branch focuses on distinguishing annotated foreground \textit{vs.} background while the other only learns what foreground objects are and thus can excavate unknown objects. The experimental results highlight the commendable performance of our approach across categories and dataset experiments in multiple metrics while maintaining a rapid detection speed.

We summarize our contributions as follows:

\begin{itemize}
    \item We widen the CAOD study to the event-based vision, which offers greater robustness against fast movement and illumination changes. To the best of our knowledge, it is the first object detection study in an open world context regarding event-based vision.
		
    \item We introduce a benchmark solution for the problem based on spatio-temporal consistency and task disentanglement, to address the core challenges of CAOD and leverage the characteristics of event cameras.
		
    \item We perform comprehensive experiments verifying the methods' efficacy. Through comparison and ablation studies, we demonstrate that the proposed DEOE outperforms the competitors across multiple evaluation settings.
\end{itemize}


\section{Related work}
\label{sec:Related}

\subsection{Event-based Object Detection}
Existing literature \cite{event_detector_2022_TIP, event_detector_2023_CVPR, event_detector_2023_TIM, event_detector_2023_AAAI} often categorizes event-based object detection into two main classes, \textit{i.e.}, those based on spiking neural networks (SNNs) \cite{event_spike_2022_CVPR, event_spike_2020_AAAI,event_spike_2021_TIP} and those utilizing dense neural network \cite{event_detector_2023_CVPR, event_detector_2022_TIP, event_fusion_2023_TPAMI}. Since the former requires efficient training methods and relies on specific hardware support, most current approaches tend to convert sparse temporal events into dense event frames for processing.

Many recent works are predominantly dedicated to exploiting the unique characteristics of event cameras while tackling their inherent limitations. For example, to match the high-speed capabilities of event cameras, research efforts such as those in \cite{event_detector_2023_CVPR, event_detector_2023_TIM} enhance the model's inference speed by streamlining model architectures and leveraging parallel computing networks. Furthermore, to make full use of the high temporal resolution features offered by event cameras and to overcome the constraint of vision sampling only during changes in brightness, other studies like \cite{event_detector_2023_CVPR, event_detector_2022_TIP, event_dataset_2020_NIPS} integrate memory networks, including Recurrent Neural Networks (RNNs) \cite{RNN_1997_TSP} and Long Short-Term Memory (LSTM) \cite{LSTM_1997_NC}, into the network architecture, which enables the model to incorporate prior information when extracting features.
	
Among the approaches mentioned, the Recurrent Vision Transformer (RVT) \cite{event_detector_2023_CVPR} achieves an optimal trade-off in accuracy and speed. Consequently, our work is built upon their introduced backbone, RVT, to obtain CAOD.

\subsection{Open-World Object Detection}
The task of Open-World Object Detection (OWOD) \cite{frame_owod_2021_CVPR, frame_owod_2022_CVPR, frame_owod_2023_CVPR} can be divided into three steps: 1) identifying objects with unknown classes in images, 2) annotating detected unknown objects manually, and 3) incrementally learning new object categories. One primary challenge in OWOD is the identification of unknown classes. Since there is no label supervision for unknown objects, distinguishing them from massive background information is a significant challenge. Towards this end, some approaches have introduced metrics such as energy scores \cite{frame_owod_2021_CVPR} and backbone activations \cite{frame_owod_2022_CVPR}, to guide discriminative decisions. In contrast to directly differentiating unknown classes from the background, Probabilistic Objectness (PROB) \cite{frame_owod_2023_CVPR} adopts a CAOD-inspired approach by first identifying general objects and then grouping them as known and unknown classes. This strategy has led to a three-fold improvement in the recall of unknown classes. 

Overall, CAOD is a fundamental step towards generic OWOD, since the class-agnostic localization of every object provides the possibility of subsequent fine classification. In this work, we take a step towards widening the application of CAOD to extreme scenarios with a new modality. Our solution is grounded in the properties of event-based vision, which distinguishes unknown objects through spatio-temporal consistency and elevates the discernment between every object and background via task disentanglement.

\begin{figure*}[!t]
    \centering
    \includegraphics[width=6.7in]{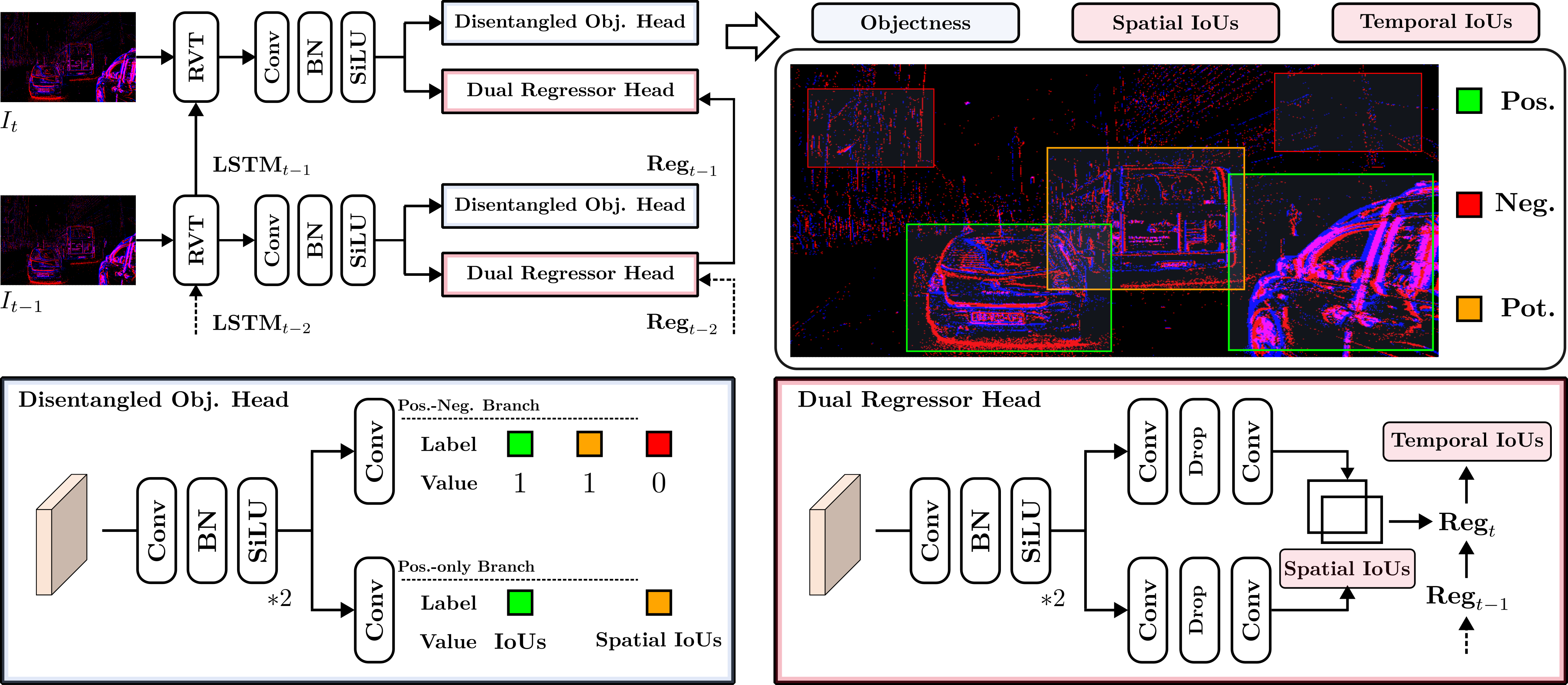}
    \caption{Approach Overview: DEOE comprises two key components: the \textit{Disentangled Objectness Head} and the \textit{Dual Regressor Head}. The former generates the ``Objectness" metric, while the latter provides the ``Spatial IoU" and ``Temporal IoU" metrics. These three metrics work in tandem to detect potential novel objects within the images, subsequently assigning positive samples based on them in the training process. The two branches in the objectness head disentangle the foreground/background division task and the foreground discovery task.}
    \label{Overview}
\end{figure*}


\section{Methodology}
\label{sec::methodology}

Before introducing the proposed DEOE, we first outline some essential foundational settings, including the event representation and the backbone. Specifically, we employ RVT \cite{event_detector_2023_CVPR} as the backbone and utilize its accompanying framing strategy for all methods in this paper to ensure a fair and equitable basis for comparison.

\textbf{Event Representation.} Asynchronous sparse event points need to be transformed into synchronous and dense event frames before they can be fed into the network. Generally, the input consists of a time interval from $t_a$ to $t_b$ discrete event points $\Phi$, with each point $e_k$ characterized by horizontal coordinate $x_k$, vertical coordinate $y_k$, polarity $p_k$, and timestamp $t_k$. To retain as much information as possible, we preserve the polarity of events, forming a 4-D tensor $E$ with shape $(2, T, H, W)$ through the equation:

\begin{align}
    \label{Event representation}
    E(p, \tau, x, y) &= \sum_{e_k \in \Phi} \delta(p - p_k) \delta(x-x_k, y - y_k) (\tau - \tau_k), \nonumber \\
    \tau_k &= \lfloor {\frac{t_k - t_a}{t_b - t_a} \cdot T} \rfloor,	 
\end{align}
where the first dimension denotes two polarities, and the second dimension denotes $T$ components associated with $T$ discretization steps. Note that to match the processing algorithm for 2D images, we reshape the tensor $E$ from $(2, T, H, W)$ to $(2T, H, W)$ before further processing. 

\textbf{Recurrent Backbone.} The backbone designed for event cameras must excel at swiftly extracting image features while seamlessly integrating temporal information, particularly in scenarios involving stationary or slow-moving objects. RVT accomplishes this by ingeniously fusing multi-axis self-attention \cite{maxvit_2022_ECCV} and LSTM \cite{LSTM_1997_NC} cells while incorporating advanced techniques like overlapping convolutions and depth-wise separable convolutions. It stands out as a remarkably proficient solution among existing approaches.

\subsection{Overview of Sample Screening}
Our sample screening scheme involves identifying potential samples ($x_{pn}$) and posing appropriate weights on them. Then, we train $x_{pn}$ together with positive samples ($x_{p}$) and negative samples ($x_{n}$). 
Specifically, we utilize confidence-based objectness $Obj$, spatial IoU $IoU_S$, and temporal IoU $IoU_T$ as selection criteria, which are grounded on the foreground confidence, spatial and temporal consistency. Specifically, we design a new score to select potential samples by sorting:

\begin{equation}
    \label{FNscore}
    Score_{p} =  Obj \cdot \sqrt{IoU_S \cdot IoU_T}.
\end{equation}

As depicted in Fig. \ref{Overview}, $IoU_S$ and $IoU_T$ are derived from the \textit{Dual Regressor Head}, while the \textit{Disentangled Objectness Head} provides $Obj$. These identified potential samples are then employed to refine the performance of these two heads, thereby enhancing their discriminative capabilities. More details will be provided in the following subsections.

\subsection{Spatio-temporal Consistency}
\label{method:dual}
The identification of potential samples is realized via spatial and temporal consistency in the \textit{Dual Regressor Head}. As shown in Fig. \ref{Overview}, unlike most previous regression heads, the \textit{Dual Regressor Head} features two regression branches, each of which includes a dropout layer to introduce a degree of randomness. We propose spatial consistency based on the following intuition: if a certain point on the feature map contains an object, then the proposals predicted by the two regression branches of the \textit{Dual Regressor Head} should exhibit significant overlap. Conversely, if it represents the background, the outputs of the two branches should show a tendency of randomness. Hence, one of our selection criteria, $IoU_S$ can be derived from the IoU between the proposals predicted by the two regression branches. To ensure that the introduction of randomness does not impact the model's performance, we mandate the \textit{Dual Regressor Head} to maximize the IoU for both $x_p$ and $x_{pn}$ by incorporating the following loss function:
\begin{equation}
    \label{spatial}
    \mathcal{L}_{sp} = -\log{(IoU_S + \varepsilon)},
\end{equation}
where $IoU_S$ exclusively encompasses $x_{p}$ and $x_{pn}$, and $\varepsilon$ represents a constant avoiding gradient explosion, set to $1\times e^{-16}$ in this work.

We propose temporal consistency based on the observation that objects tend to re-occur for adjacent time stamps in the event stream. In contrast to traditional cameras, event cameras can capture a frame of events in an extremely short time interval (close to 1 ms), which leads to objects having nearly identical positions in consecutive frames. Therefore, we define the $IoU_T$ as the IoU between proposals corresponding to the same feature points in the same position across consecutive frames. If the predicted proposals at adjacent time $t$ and $t-1$ have a high $IoU_T$, they are more likely to contain an object. Conversely, if both proposals at time $t$ and $t-1$ represent the background, their positions should exhibit some randomness, resulting in a lower $IoU_T$. Both $IoU_S$ and $IoU_T$ serve as selection criteria, working in the identification of potential samples.

The outputs from the two branches are averaged to obtain the final regression box, and the IoU loss is computed based on the overlap between the predicted and the GT boxes:
\begin{equation}
    \label{iou}
    \mathcal{L}_{iou} = 1 - \frac{|\mathcal{R}(x_p) \cap \text{GT}|}{|\mathcal{R}(x_p) \cup \text{GT}|},
\end{equation}
where $\mathcal{R}(\cdot)$ denotes the predicted bounding boxes and GT denotes the GT bounding boxes. 

\subsection{Task Disentanglement}
\label{method:disentangled}
Prior research \cite{frame_caod_2021_RAL, frame_caod_2022_ECCV, frame_owod_2023_CVPR} has emphasized a limitation in applying standard detection methods to CAOD, namely the substantial penalty for objects without annotations, leading to insufficient supervision towards potential objects. Conversely, when employing solely positive samples and incorporating soft labels \cite{soft_label_2018_ECCV, frame_caod_2021_RAL}, it can fail to distinguish between objects and backgrounds. To address this concern, we propose a \textit{Disentangled Objectness Head} to decouple the division of foreground and background and the discovery of unknown objects, consisting of two branches: the positive-negative branch and the positive-only branch.

The positive-negative branch is employed for the concurrent training of $x_p$, $x_{pn}$, and $x_n$. For the $x_p$ and $x_{pn}$, labels are set to 1, while $x_n$ are labeled as 0. The training loss of the positive-negative branch is given by:
\begin{equation}
    \label{losspn}
    \resizebox{.9\hsize}{!}{$\mathcal{L}_{pn} = BCE(\mathcal{O}_{pn}(x_p), 1) + wBCE(\mathcal{O}_{pn}(x_{pn}), 1) + BCE(\mathcal{O}_{pn}(x_n), 0)$},
\end{equation}
where $\mathcal{O}_{pn}(\cdot)$ stands for the output by the positive-negative branch. Considering that potential samples identified by the proposed clue cannot be entirely accurate, they should not be assigned the same weight as GT samples. Instead, each potential sample in Eq. \ref{losspn} is assigned a weight denoted as $w$, which is determined based on the $Obj$ confidence from the \textit{Disentangled Objectness Head} and the $IoU_S$ from the \textit{Dual Regressor Head}, \textit{i.e.}, the specific formula for calculating $w$ is as follows:
\begin{equation}
    \label{weight}
    w = \sqrt{Obj \cdot{IoU_S}},
\end{equation}
where both $Obj$ and $IoU_S$ are in the range of $[0, 1]$, leading to the confinement of $w$ within this interval, guaranteeing that potential samples are weighted lower than GT samples. Furthermore, we adopt a weighting scheme following the method in \cite{prime_attention_2020_CVPR} to keep the loss value unchanged when multiplied by the assigned weights. 

Differently, for the positive-only branch, negative samples $x_n$ are no longer suppressed in the training process. By doing so, we would like to empower this branch with the capability to maintain confidence in unknown objects. Labels for $x_p$ are determined based on the IoU between anchor boxes and the GT boxes $IoU_G$, whereas labels for $x_{pn}$ are derived from the $IoU_S$ output by the \textit{Dual Regressor Head}. In short, for the positive-only branch, the loss can be expressed as:
\begin{equation}
    \label{losspo}
    \resizebox{.9\hsize}{!}{$\mathcal{L}_{po} = BCE(\mathcal{O}_{po}(x_p), IoU_G) + BCE(\mathcal{O}_{po}(x_{pn}), IoU_S)$},
\end{equation}
where $\mathcal{O}_{po}(\cdot)$ stands for the output by the positive-only branch. In conclusion, the total loss for model training can be expressed as:
\begin{equation}
    \label{loss}
    \mathcal{L} = \mathcal{L}_{pn} + \mathcal{L}_{po} + \mathcal{L}_{sp} + \mathcal{L}_{iou}.
\end{equation}

Besides, we set the potential sample count as an adjustable parameter, and we assess its impact on the model's performance in the experimental section.
 

\subsection{Model Inference}
Different from previous detection models, each of the proposed heads in DEOE features two branches. Thus, the final inference combines predictions from these dual branches. Specifically, the $Obj$, serving as an indicator of objectness, is derived by multiplying the outputs of the positive-negative branch and positive-only branch in the \textit{Disentangled Objectness Head}. Likewise, we obtain box predictions by averaging the outputs from the two branches in the \textit{Dual Regressor Head}. The ultimate result is derived from the outputs of two heads, forming a set of five elements $\{Obj, x, y, w, h\}$. 
\section{Experiments}
\label{sec:experiments}

\subsection{Experimental Setup}
\textbf{Dataset.} We use the 1 MEGAPIXEL Automotive Detection Dataset (Mpx) \cite{event_dataset_2020_NIPS} dataset as our primary experimental dataset, considering its large scale and class number over other existing event camera datasets. This dataset not only contains close to 25 million annotated bounding boxes but also features 7 distinct detection categories (0: pedestrians, 1: two-wheelers, 2: cars, 3: trucks, 4: buses, 5: traffic signs, 6: traffic lights). Moreover, these images boast a high resolution of $1280 \times 720$ and are annotated at frequencies of 30 or 60 Hz, rendering it the largest event camera object detection dataset currently available. 
	
However, the 1 Mpx dataset's annotations are pseudo-labels generated by the image-based model and have not undergone manual refinement, resulting in a higher likelihood of errors, particularly for categories other than pedestrians, two-wheelers, and cars. Additionally, due to its substantial data volume, training on this dataset is quite time-consuming. Considering these factors, we sample approximately 4 hours worth of high-confidence data for evaluation according to the confidence scores by sorting, which accounts for roughly a quarter of the whole dataset. This strategy not only enhances the reliability of evaluation but also reduces the complexity of reproducing our results. Despite using a subset of the 1 Mpx data, it still comprises 7.5 million annotated bounding boxes, which surpass all other existing event-based detection datasets, with class distribution detailed in Tab. \ref{class}.

\begin{table}[!t]
    \renewcommand{\arraystretch}{1.2} 
    \centering
    \caption{Class distribution of Four Hour 1 Mpx dataset.}
    \setlength{\tabcolsep}{3.5pt} 
    \vspace{-6pt}
    \label{class}
    \begin{tabular}{ccccccccc}
        \hline
        Dataset & 0 & 1 & 2 & 3 & 4 & 5 & 6 & Confidence \\
        \hline
        Train & 360k & 69k & 2364k & 277k & 33k & 318k & 169k & 0.973 \\
        
        Val & 410k & 52k &	1152k &	100k & 26k & 201k &	154k & 0.963 \\
        
        Test & 139k& 41k & 1213k & 109k & 44k & 191k & 86k & 0.971 \\
        
        \hline
    \end{tabular} 
\end{table}

Some previous research \cite{event_detector_2023_TIM, event_detector_2023_AAAI} also employs the subsets of 1 Mpx due to its inaccurate annotations and extensive dataset sizes. For instance, Liu \emph{et al.}  \cite{event_detector_2023_TIM} employed a downsampling technique to reduce the annotation frequency from 60 Hz to 1 Hz, reducing the data volume by a factor of 60. However, this process disrupted the temporal continuity to some extent, making it less compatible with our proposed approach. On the other hand, Wang \emph{et al.} \cite{event_detector_2023_AAAI} mitigated annotation inaccuracies by manually annotating a subset of the dataset. Yet, the subset dataset only offered annotations for three categories, which is inappropriate for the CAOD task.

To further evaluate the model's generalization, we conducted a cross-dataset evaluation on the DSEC-Detection dataset \cite{dataset_2021_RAL, dsec_det_2022_arxiv}. This dataset comprises 60 sequences, totaling approximately 390k annotated bounding boxes across 8 categories.

\textbf{Evaluation Protocol.} To simulate an open-world object detection scenario, we partition the categories, designating a portion as known classes for training and the remaining as unknown classes to be discovered. Specifically, we define two tasks distinguished by ``known+unknown" classes, \textit{i.e.}, ``5+2" and ``4+3" settings, as detailed in Tab. \ref{task}. Note that in both tasks, we have designated pedestrians, cars, traffic signs, and traffic lights as known classes. This decision is driven by the common practice of annotating the ``pedestrians" and ``cars" in the majority of autonomous driving datasets. However, the ``traffic signs" and ``traffic lights", owing to the small size of the targets, often have inaccurate annotations in pseudo-labels. To bolster the reliability of positional assessments, we include them as known samples exclusively for training purposes.
Building upon this configuration, we train with images containing known classes and subsequently evaluate images containing unknown classes. 
In our experiments, we report the model's detection performance on both known classes and unknown classes to better understand the model's behaviour in diverse scenarios.  

Following the existing CAOD works \cite{frame_caod_2021_RAL, frame_caod_2022_ECCV, owod_seg_2021_CVPR}, we use the Average Recall (AR) as the main metric. In addition, we employ the area under the curve (AUC) proposed in \cite{frame_caod_2021_RAL} for further evaluation. Note that the Average Precision (AP) is not commonly used for assessing CAOD, considering that the main aim of CAOD is to discover unknown objects, and the inadequacy in data annotation will introduce inconsistencies when utilizing mAP for evaluating CAOD tasks.

\textbf{Implementation Details.} We employ RVT-Base \cite{event_detector_2023_CVPR} as our backbone, opting for a training approach starting from scratch without using any pre-trained models. For the sake of memory efficiency and speed improvement, we set the sequence length as 5 and implement a $2\times$ down-sampling to the images, resulting in images of size $640 \times 360$, which follows the common practice in existing research \cite{event_detector_2023_CVPR, event_detector_2022_TIP, event_dataset_2020_NIPS}. The training batch size is set to 8, with a maximum learning rate of $2 \times 10^{-4}$, and optimization is performed using the Adam optimizer. All experiments are conducted on a single RTX4090 GPU. Each training session comprised 200,000 iterations, and the total training duration is approximately 30 hours.

 \begin{table}[!t]
    \renewcommand{\arraystretch}{1.2} 
    \centering
    \caption{The detailed configuration of closed-set and open-set classes.}
    \setlength{\tabcolsep}{3.5pt} 
    \vspace{-6pt}
    \label{task}
    \begin{tabular}{c|cc|cc}
        \hline
        \multirow{2}{*}{classes} & \multicolumn{2}{c}{$\text{Task}_1$} & \multicolumn{2}{c}{$\text{Task}_2$}  \\
        \cline{2-5}
        & Known  & Unknown  & Known  & Unknown  \\
        \hline
        pedestrians    & \checkmark  &            & \checkmark &		     \\
        
        two-wheelers   &             & \checkmark &            & \checkmark  \\
        
        cars 		   & \checkmark  &     	 	  & \checkmark &  			 \\
        
        trucks 		   & \checkmark  &  		  & 		   & \checkmark  \\
        
        buses 		   &   			 & \checkmark &            & \checkmark  \\
        
        traffic signs  & \checkmark  & 			  & \checkmark &   			 \\
        
        traffic lights & \checkmark  & 			  & \checkmark &   			 \\
        
        \hline
    \end{tabular} 
\end{table}

\subsection{Comparison with the baselines}
\subsubsection{Baselines}
Given the absence of open-set studies in the field of event-based object detection, we develop the following baselines for comparison by incorporating novel ideas from existing RGB-based open-set detection research, and these baselines are built upon the state-of-the-art event-based object detection model RVT:

\begin{itemize}
    \item $\text{CA-RVT}$. We discard the classification head in the default RVT-Base model \cite{event_detector_2023_CVPR}, leaving it to only output object scores, and it serves as the fundamental baseline.
    
    \item $\text{CA-RVT}_\text{O}$. Following \cite{frame_owod_2021_CVPR}, we shift from training the model with binary labels to employing localization scores (IoU) for positive samples. 
    
    \item $\text{CA-RVT}_\text{P}$. Following the idea of work \cite{frame_owod_2021_CVPR}, we select a
    part of negative samples with higher objectness scores and assign them as potential samples for training.
\end{itemize}

Apart from the aforementioned scenario, we also introduce an ideal condition where the model can identify and include all potential samples during training. We refer to this condition as the \textbf{Oracle} setting.

\begin{table*}[!t]
    \renewcommand{\arraystretch}{1.2} 
    \centering
    \caption{\textbf{Main results on the Four Hour 1 Mpx dataset under two settings.} Note that the ``$\text{AR}_\text{k}$" denotes the average recall when the maximum number of detection boxes is set to $\text{k}$. Additionally, when calculating the $\text{AR}$ for unknown classes, we exclude the boxes that enclose known samples from the count within $\text{k}$.}
    \setlength{\tabcolsep}{3.5pt} 
    \vspace{-6pt}
    \label{results}
    \begin{tabular}{c|c|cccccc|cccccc|c}
        \hline
        \multirow{7}{*}{5+2 setting}  & \multirow{2}{*}{Methods} & \multicolumn{6}{c|}{Unknown} & \multicolumn{6}{c|}{All (+ two-wheelers, trucks)} & \multirow{2}{*}{Time (ms)}  \\
        \cline{3-14}
        & & $\text{AUC}$ & $\text{AR}_{10}$ & $\text{AR}_{30}$ & $\text{AR}_{50}$ & $\text{AR}_{100}$ & $\text{AR}_{300}$ & $\text{AUC}$ & $\text{AR}_{10}$ & $\text{AR}_{30}$ & $\text{AR}_{50}$ & $\text{AR}_{100}$ & $\text{AR}_{300}$&\\
        \cline{2-15}
        \rowcolor{green!20}
        \cellcolor{white} &  Oracle & $27.78$ & $33.90$ &$34.77$ & $35.14$ & $35.57$ & $36.03$ & $36.81$ & $43.99$ & $46.60$ & $47.13$& $47.74$ & $48.21$ & $6.52$\\	
        & $\text{CA-RVT}$ & $17.77$ & $19.15$ & $22.49$ & $23.71$ & $24.85$ & $25.74$ & $34.88$ & $41.38$ & $44.10$ & $44.83$& $45.52$ & $46.10$ & $6.19$ \\
        & $\text{CA-RVT}_\text{O}$ & $19.79$ & $22.01$ & $24.85$ & $25.97$ & $27.25$ & $28.06$ & $35.44$ & $\bm{41.54}$ & $44.89$ & $45.71$& $46.67$ & $47.39$ & $6.27$ \\	
        & $\text{CA-RVT}_\text{P}$ & $19.19$ & $21.86$ & $24.02$ & $24.77$ & $25.89$ & $27.00$ & $34.69$ & $41.14$ & $43.96$ & $44.54$& $45.17$ & $45.91$ & $6.07$ \\	
        & $\text{DEOE (ours)}$ & $\bm{22.70}$ & $\bm{23.69}$ & $\bm{28.57}$ & $\bm{30.10}$ & $\bm{32.02}$ & $\bm{35.28}$ & $\bm{36.11}$ & $41.35$ & $\bm{45.64}$ & $\bm{46.83}$& $\bm{48.22}$ & $\bm{50.08}$ & $8.82$ \\
        \hline	
        \multirow{7}{*}{4+3 setting}  & \multirow{2}{*}{Methods} & \multicolumn{6}{c|}{Unknown} & \multicolumn{6}{c|}{All (+ two-wheelers, buses, trucks)} & \multirow{2}{*}{Time (ms)}   \\
        \cline{3-14}
        & & $\text{AUC}$ & $\text{AR}_{10}$ & $\text{AR}_{30}$ & $\text{AR}_{50}$ & $\text{AR}_{100}$ & $\text{AR}_{300}$ & $\text{AUC}$ & $\text{AR}_{10}$ & $\text{AR}_{30}$ & $\text{AR}_{50}$ & $\text{AR}_{100}$ & $\text{AR}_{300}$&\\
        \cline{2-15}
        \rowcolor{green!20}
        \cellcolor{white} & Oracle & $35.17$ & $43.09$ & $44.06$ & $44.38$ & $44.87$ & $45.45$ & $40.63$ &  $49.16$ & $51.30$& $51.69$ & $52.19$ & $52.67$& $6.42$ \\	
        & $\text{CA-RVT}$& $8.83$ & $8.61$ & $11.22$ & $12.09$ & $13.15$ & $13.99$ & $34.61$ & $41.36$ & $43.78$& $44.28$ & $44.89$ & $45.41$ & $6.36$ \\
        & $\text{CA-RVT}_\text{O}$ & $8.97$ & $8.74$ & $11.16$ & $12.21$ & $13.47$ & $14.40$ & $34.59$ & $41.37$& $43.57$ & $44.21$ & $44.91$ & $45.53$ & $6.47$ \\	
        & $\text{CA-RVT}_\text{P}$ & $9.00$ & $8.90$ & $11.45$ & $11.98$ & $13.20$ & $14.44$ & $34.94$ & $\bm{41.94}$& $44.09$ & $44.59$ & $45.18$ & $45.77$ & $6.19$ \\	
        & $\text{DEOE (ours)}$ & $\bm{12.49}$ & $\bm{11.44}$ & $\bm{14.69}$ & $\bm{16.48}$ & $\bm{19.11}$ & $\bm{23.48}$ & $\bm{35.95}$ & $41.66$& $\bm{45.08}$ & $\bm{46.27}$ & $\bm{47.73}$ & $\bm{49.70}$ & $8.91$ \\	
        \hline
    \end{tabular} 
\end{table*}

\subsubsection{Main Results}
\textbf{Effectiveness.} As shown in Table \ref{results}, the performance of our DEOE significantly surpasses different baselines under both unknown class and all class settings. Notably, DEOE even outperforms our designated \textbf{Oracle} on several metrics, \textit{i.e.}, $\text{AR}_{100}$ and $\text{AR}_{300}$ under ``5+2" setting. This improvement can be attributed to two factors. Firstly, the DEOE model structure contributes to the observed gain. Secondly, the potential samples selected by DEOE may include samples whose categories are not present in the GT classes. By training these potential samples as positive samples, the generalization ability of DEOE is further enhanced. 

\textbf{Efficiency.} We also evaluated the inference time on the Four Hour 1 Mpx dataset (from input to prediction output) with the batch size of 1, and the results are presented in the right column of Table \ref{results}. Since both DEOE and the comparable baselines leverage the currently fast backbone RVT in a one-stage detection pipeline, all the methods consistently achieve over \textbf{100 FPS}. In addition, the event-based method does not require additional time for imaging (approximately 30 ms) \cite{event_detector_2023_CVPR}, instead, it only needs a frame construction time close to 1 ms.

\begin{table}[!t]
    \renewcommand{\arraystretch}{1.2} 
    \centering
    \caption{\textbf{Ablations of Detection Heads.} The contributions of the \textit{Disentangled Objectness Head} and the \textit{Dual Regressor Head}, respectively.}
    \setlength{\tabcolsep}{2.8pt} 
    \vspace{-6pt}
    \label{branchs}
    \begin{tabular}{c|c|ccc|ccc}
        \hline
        \multirow{2}{*}{Disent.} & \multirow{2}{*}{Dual. (15)} & \multicolumn{3}{c|}{Unknown}
        & \multicolumn{3}{c}{All} \\
        \cline{3-8}
        &  & $\text{AUC}$ & $\text{AR}_{30}$ & $\text{AR}_{300}$ & $\text{AUC}$ & $\text{AR}_{30}$ & $\text{AR}_{300}$ \\
        \hline
        &  & $17.77$ & $22.49$ & $25.74$ & $34.88$ & $44.10$ & $46.10$ \\
        \checkmark &  & $20.79$ & $26.00$ & $30.73$ & $35.50$ & $44.81$ & $48.34$ \\
        & \checkmark  & $20.29$ & $24.80$ & $32.27$ & $33.83$ & $42.81$ & $46.77$  \\
        \checkmark & \checkmark & $\bm{22.70}$&$\bm{28.57}$ & $\bm{35.28}$ & $\bm{36.11}$ & $\bm{45.64}$ &$\bm{50.08}$\\
        
        \hline
    \end{tabular} 
\end{table}

\subsection{Ablation Studies}
In this section, we delve into the impact of the two pivotal components of DEOE, the \textit{Disentangled Objectness Head} and \textit{Dual Regressor Head}, on the final performance of the model. Furthermore, we conduct ablation experiments on key structures and hyper-parameters within each branch, which encompasses ablating the branch structure within the \textit{Disentangled Objectness Head} and changing the number of potential samples in the \textit{Dual Regressor Head}. Note that all the ablations are conducted under the ``5+2" setting.

\subsubsection{Detection Head}
The contributions of the \textit{Disentangled Objectness Head} and the \textit{Dual Regressor Head} are presented in Tab. \ref{branchs}, respectively. Both components play a crucial role in significantly enhancing the capabilities of the model. The benefits introduced by the \textit{Disentangled Objectness Head} are particularly noticeable in unknown class samples and across all-class samples, verifying the advantages of task disentanglement. Conversely, the \textit{Dual Regressor Head} exhibits a moderate reduction in the known class recall due to the introduction of potential samples. However, the synergy of the two heads rectifies this limitation, leading to a superior enhancement.

\subsubsection{Task Disentanglement}

\begin{table}[!t]
    \renewcommand{\arraystretch}{1.2} 
    \centering
    \caption{\textbf{Disentangled Objectness Head.} The impact of different branch combinations on its performance.}
    \setlength{\tabcolsep}{3.5pt} 
    \vspace{-6pt}
    \label{branch}
    \begin{tabular}{c|c|cc|ccc}
        \hline
        \multirow{2}{*}{Pos. \& Neg.} & \multirow{2}{*}{Pos. Only} & \multicolumn{2}{c|}{Unknown}
        & \multicolumn{3}{c}{All} \\
        \cline{3-7}
        &  & $\text{AR}_{30}$ & $\text{AR}_{100}$ & $\text{AR}_{10}$ & $\text{AR}_{30}$ & $\text{AR}_{100}$ \\
        \hline
        \checkmark &  & $22.49$ & $24.85$ & $\bm{41.38}$ & $44.10$ & $45.52$ \\
      & \checkmark  & $23.28$ & $\bm{29.56}$ & $33.79$ & $41.53$ & $\bm{48.84}$ \\
        \checkmark	  & \checkmark  & $\bm{26.00}$ & $29.13$ & $41.10$ & $\bm{44.81}$ & $47.19$ \\
        \hline
    \end{tabular} 
\end{table}

We analyze the impact of the positive-negative branch and positive-only branch within the \textit{Disentangled Objectness Head} in Tab. \ref{branch}. When exclusively utilizing the positive-negative branch, the model exhibits a decreased recall for unknown class samples due to its suppression of unknown objects. In contrast, employing solely the positive-only branch enhances recall specifically for unknown classes. However, its over-representation of background regions leads to decreased recall for known classes, especially when the maximum number of detection boxes is limited. 

Nevertheless, when these two branches synergize, they concurrently retain the suppressive effect on the background provided by the positive-negative branch and the generalization capability of the positive-only branch. The task disentanglement and collaboration culminate in a substantial enhancement in recall for both unknown classes and all class samples. Note that, for a purer comparative analysis, none of the experiments conducted here employed the \textit{Dual Regressor Head}.

\begin{table}[!t]
    \renewcommand{\arraystretch}{1.2} 
    \centering
    \caption{\textbf{Dual Regressor Head.} The influence of the number of potential samples on the model.}
    \setlength{\tabcolsep}{3.5pt} 
    \vspace{-6pt}
    \label{FNsample}
    \begin{tabular}{c|ccc|ccc}
        \hline 
        \multirow{2}{*}{Pot. Count}  & \multicolumn{3}{c|}{Unknown} &  \multicolumn{3}{c}{All} \\ 
        \cline{2-7}
        & $\text{AUC}$ & $\text{AR}_{30}$ & $\text{AR}_{300}$ & $\text{AUC}$ & $\text{AR}_{30}$ & $\text{AR}_{300}$ \\
        \hline
        $0$   & $20.25$ & $26.19$ & $30.46$ & $34.55$ & $43.68$ & $48.33$  \\
        $5$   & $21.51$ & $26.76$ & $35.09$ & $34.82$ & $44.03$ & $49.10$  \\
        $15$  & $22.48$ & $27.89$ & $\bm{35.82}$ & $36.01$ & $45.51$ & $\bm{50.22}$  \\
        $25$  & $21.98$ & $26.75$ & $35.29$ & $35.28$ & $44.40$ & $49.97$  \\ 
        $35$  & $\bm{22.70}$ & $\bm{28.57}$ & $35.28$ & $\bm{36.11}$ & $\bm{45.64}$ & $50.08$  \\ 
        $75$  & $19.62$ & $24.75$ & $29.57$ & $35.04$ & $44.27$ & $47.88$  \\ 
        $100$ & $20.41$ & $25.47$ & $30.47$ & $34.57$ & $43.56$ & $47.20$  \\ 
        \hline
    \end{tabular}
\end{table}

\subsubsection{Potential Sample Count}
We investigate the influence of the selected number of potential samples, presented in Tab. \ref{FNsample}. From this table, we can observe that both excessively small and excessively large numbers of potential samples result in a performance decline for DEOE. The reason for the deterioration with an overly small number is that the model cannot adequately learn from the information contained in potential samples. In contrast, if the number of potential samples is set too large, it will inevitably lead to a degradation in the quality of the positive samples. In other words, the selected potential samples may not accurately represent unknown objects, when the model erroneously treats these incorrectly labeled samples as positive samples during training, it will result in a decline in performance. 

\subsection{Cross-dataset Validation}
We employ the DSEC-Detection dataset for cross-dataset validation. Specifically, 
We utilize Four Hour 1 Mpx for training and evaluate the model on the DSEC-Detection dataset. In contrast to previous assessments, we restrict the training classes to only two classes, \textit{i.e.}, pedestrians and cars, which are the most prevalent in the autonomous driving dataset. During the evaluation, we extend the assessment on all classes present in the DSEC-Detection dataset.

Tab. \ref{cro-results} presents the quantitative results of the cross-dataset experiments. Owing to the limited number of known classes, the baseline model exhibits poor performance on unknown classes. In contrast, the proposed DEOE, equipped with the ability to automatically explore potential samples, shows commendable performance even in scenarios with minimal samples from known classes.

It is noteworthy that in the baseline approach, $\text{CA-RVT}_\text{P}$ similarly incorporates a subset of potential samples for training. However, due to its singular discriminative criterion, these potential samples do not contribute substantially to the improvement of performance beyond that achieved by the $\text{CA-RVT}$ baseline.

\subsection{Visual Results}

\begin{figure*}[!t]
    \centering
    \includegraphics[width=6.9in]{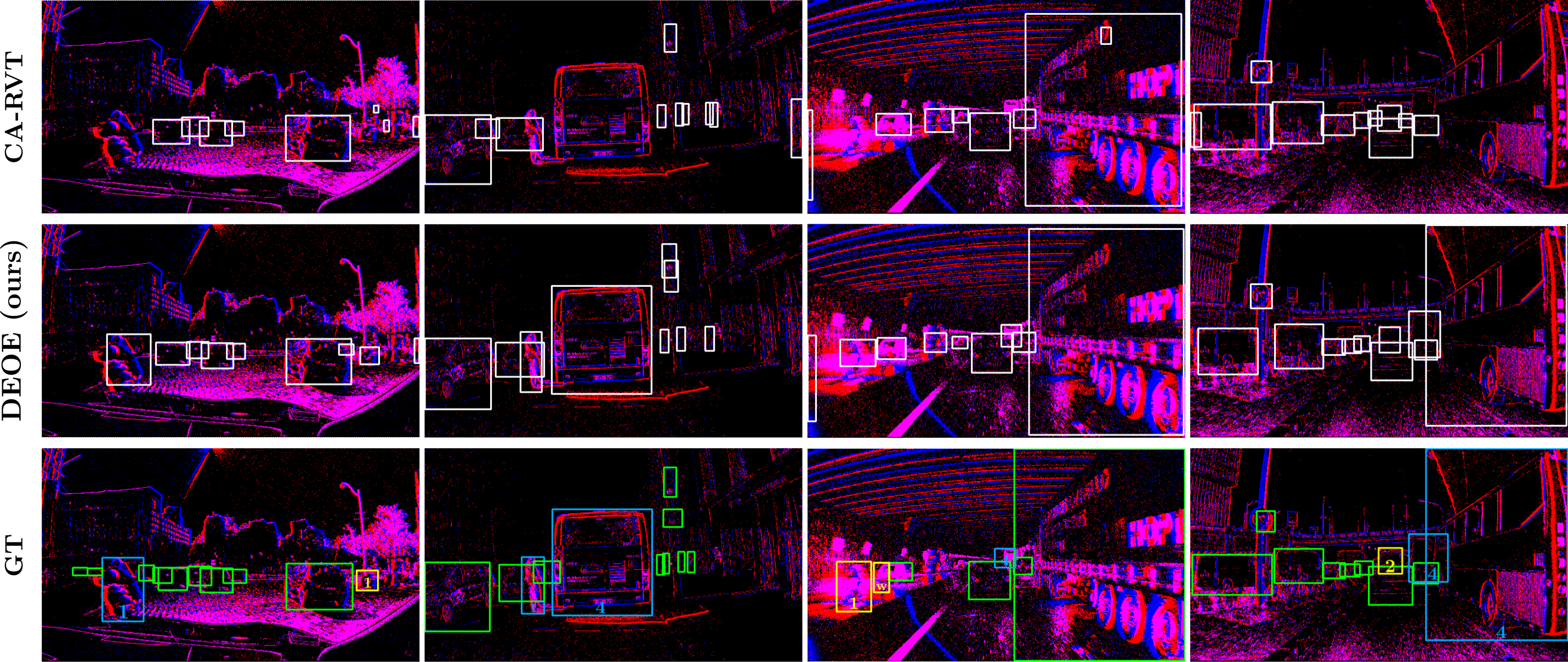}
    \caption{Qualitative results on example images from Four Hour 1 Mpx dataset (5+2 setting). upper: the detection results of CA-RVT; middle: the detection results of DEOE; bottom: the GT annotations. Note that in GT annotations, three different colored bounding boxes are employed, the \textcolor{green}{\textbf{Green Boxes}} signify annotations for known classes (people, cars, and so on), \textcolor{myblue}{\textbf{Blue Boxes}} designate annotations for unknown classes (two-wheelers and bus), \textcolor{myyellow}{\textbf{Yellow Boxes}} represent conspicuously wrong annotations including false alarms (signified by "W") and missed detection, while the white bounding boxes in the first two rows represent the prediction results of the two models. The number at the bottom of the box indicates the category of the detected object.}
    \label{vis}
\end{figure*}

            
            
            

\begin{figure}[!t]
    \centering
    \subfloat[$\footnotesize \text{Event pred.}$]{
        \begin{minipage}[b]{0.49\linewidth}
            \includegraphics[width=1\linewidth]{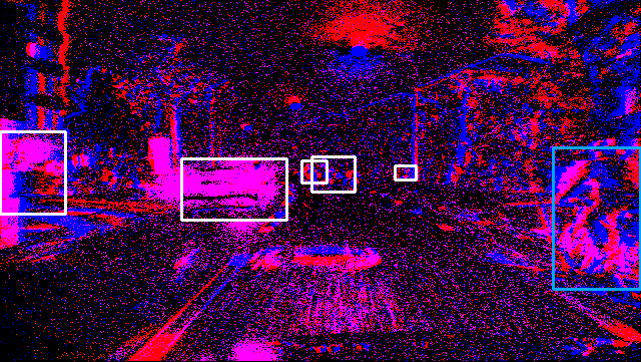}\vspace{1mm}
            \includegraphics[width=1\linewidth]{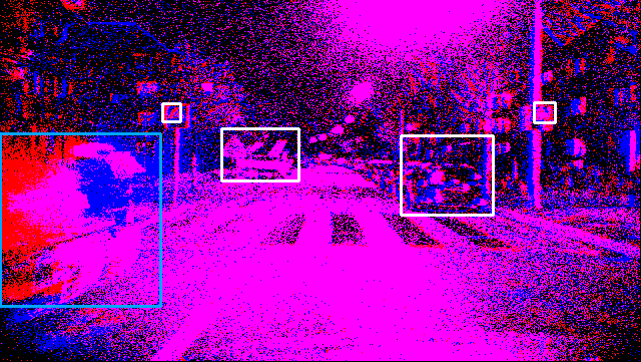}\vspace{1mm}
    \end{minipage}}
    \subfloat[$\footnotesize \text{Image pred.}$]{
        \begin{minipage}[b]{0.49\linewidth}
            \includegraphics[width=1\linewidth]{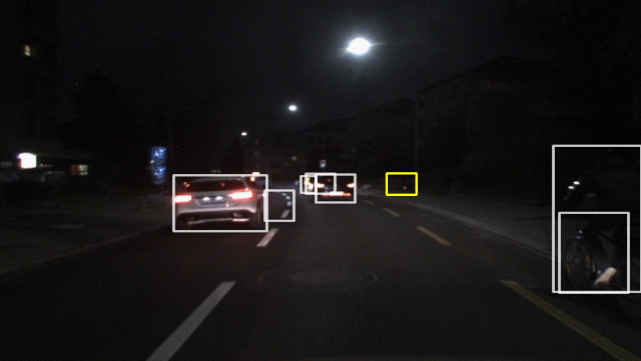}\vspace{1mm}
            \includegraphics[width=1\linewidth]{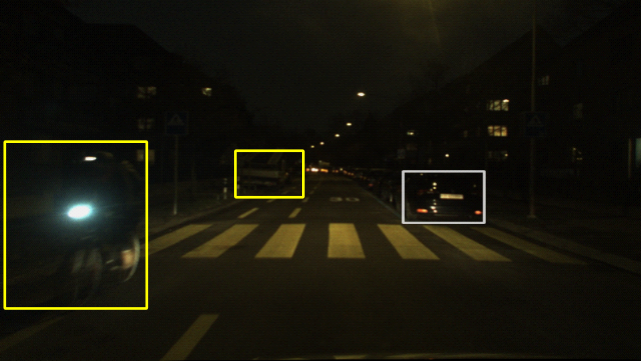}\vspace{1mm}
    \end{minipage}}

    \caption{A comparison between event cameras and RGB cameras in extreme scenarios. The white boxes in the image denote the model's detection results for known classes, the \textcolor{myblue}{\textbf{Blue Boxes}} designate unknown classes, and the \textcolor{myyellow}{\textbf{Yellow Boxes}} represent instances where the image-based model missed detections compared to the event-based model. Note that the model corresponding to event prediction came from the previous ``5+2'' setting task, while the model for image prediction is a pretrained YOLOX model derived from mmdetection.}
    \label{vis2}
\end{figure}

The predictions in Fig. \ref{vis} illustrate a comparison between DEOE and CA-RVT. While CA-RVT performs well on known classes such as pedestrians and cars, it falls short in identifying unknown classes like two-wheelers and buses. In contrast, DEOE addresses this limitation of CA-RVT and can even discover objects that are missed in the GT annotations.

We also visualize predictions under extreme conditions, as depicted in Fig. \ref{vis2}. We can find that RGB-based methods struggle under extreme light conditions and with fast-moving objects. In the first row, the RGB-based method misses a car in the shadowed area. Worsely, in the second row, the RGB-based method can only detect a car in the light, while overlooking a swiftly moving two-wheeler and a cargo-laden vehicle. By contrast, the event-based method fares well in both scenarios.

\begin{table*}[!t]
    \renewcommand{\arraystretch}{1.2} 
    \centering
    \caption{\textbf{Cross dataset validation.} In this experiment, we train on the Four Hour 1 Mpx dataset and evaluate on the DSEC-Detection dataset to verify DEOE's generalization ability.}
    \setlength{\tabcolsep}{3.5pt} 
    \vspace{-6pt}
    \label{cro-results}
    
        \begin{tabular}{c|ccccccc|cccccc|c}
            \hline
            \multirow{2}{*}{Methods} & \multicolumn{7}{c|}{Unknown} & \multicolumn{6}{c|}{All (+ pedestrians, cars)} & \multirow{2}{*}{Time (ms)} \\
            \cline{2-14}
            & $\text{AP}$ &$\text{AUC}$& $\text{AR}_{10}$ & $\text{AR}_{30}$ & $\text{AR}_{50}$ & $\text{AR}_{100}$ & $\text{AR}_{300}$ & $\text{AUC}$ & $\text{AR}_{10}$ & $\text{AR}_{30}$ & $\text{AR}_{50}$ & $\text{AR}_{100}$ & $\text{AR}_{300}$& \\
            \cline{1-15}
            $\text{CA-RVT}$ &$1.32$& $10.27$ & $10.60$ & $12.82$ & $13.79$ & $14.79$ & $15.81$ & $20.37$ & $23.54$ & $25.73$ & $26.42$& $27.11$ & $27.76$ & $6.21$ \\
            $\text{CA-RVT}_\text{O}$& $1.42$ & $10.63$ & $10.66$ & $13.09$ & $14.09$ & $15.64$ & $17.24$ & $21.58$ &  $24.35$ & $27.35$ & $28.18$& $29.15$ & $30.21$ & $6.12$ \\	
            $\text{CA-RVT}_\text{P}$ & $1.62$& $11.16$ & $11.23$ & $14.10$ & $15.11$ & $16.27$ & $17.36$ & $21.32$ & $\bm{24.78}$ &$26.92$ & $27.55$& $28.28$ & $28.94$&$6.48$ \\	
            $\text{DEOE (ours)}$&$\bm{3.22}$ & $\bm{15.07}$ & $\bm{13.60}$ & $\bm{17.99}$ & $\bm{20.14}$ & $\bm{23.23}$ & $\bm{28.02}$ & $\bm{23.25}$ & $24.66$ & $\bm{28.84}$ & $\bm{30.70}$& $\bm{32.87}$ & $\bm{35.57}$&$8.75$ \\
            \hline
        \end{tabular} 
\end{table*}

\begin{figure}[!t]
    \centering
    \subfloat[$\footnotesize \text{RGB Image}$\label{vis3:ori}]{
        \begin{minipage}[b]{0.32\linewidth}
            \includegraphics[width=1\linewidth]{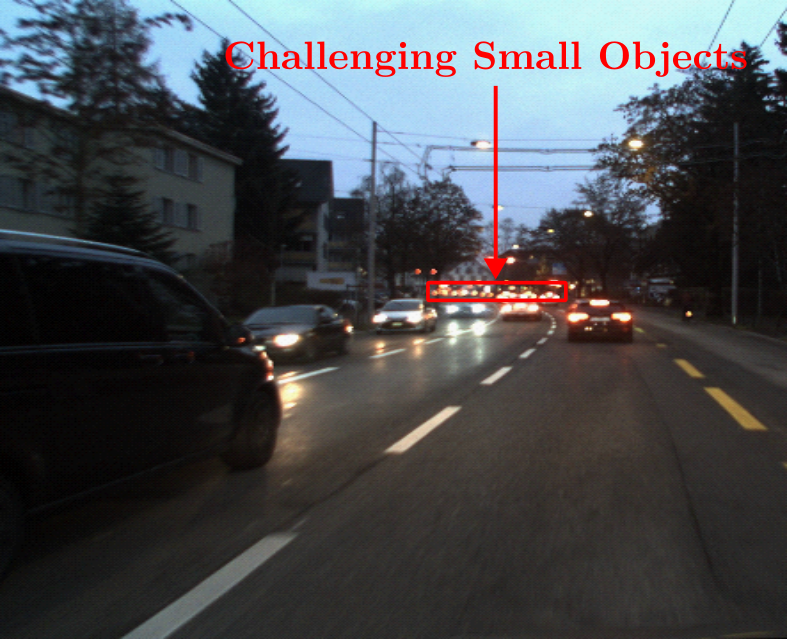}\vspace{1mm}
        \end{minipage}}
    \subfloat[$\footnotesize \text{DEOE}$\label{vis3:event}]{
        \begin{minipage}[b]{0.32\linewidth}
            \includegraphics[width=1\linewidth]{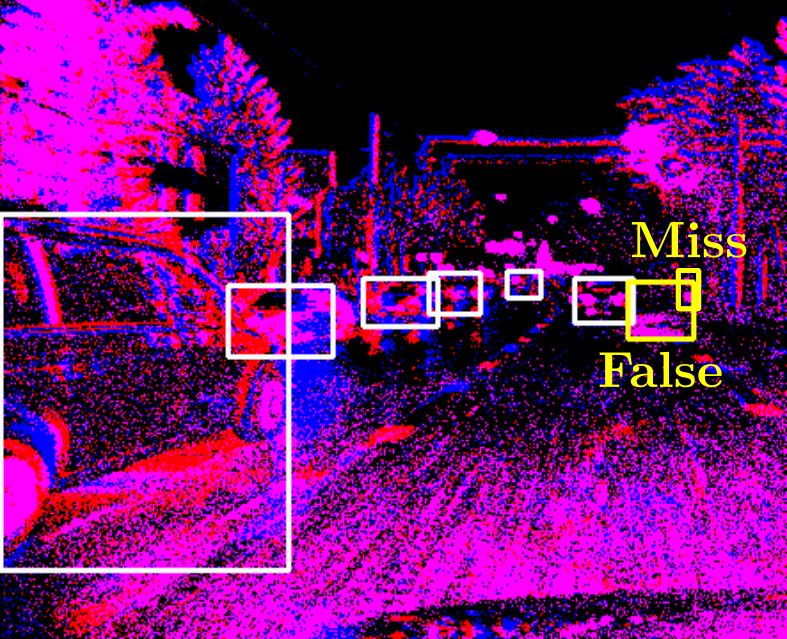}\vspace{1mm}
        \end{minipage}}
    \subfloat[$\footnotesize \text{GT}$\label{vis3:pseGT}]{
        \begin{minipage}[b]{0.32\linewidth}
            \includegraphics[width=1\linewidth]{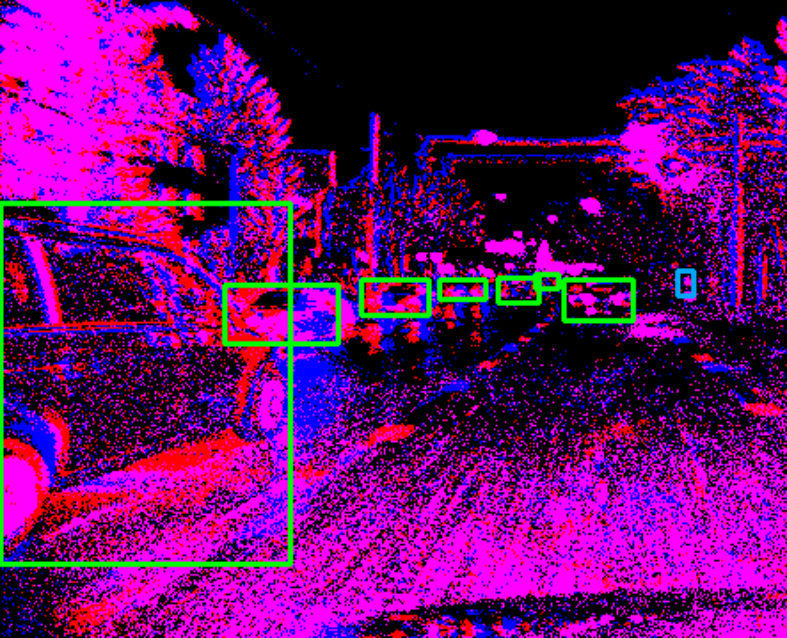}\vspace{1mm}
        \end{minipage}}
    \caption{Some typical failure cases of DEOE on DSEC-Detection dataset. The \textcolor{red}{Red Boxes} contain challenging small objects, positioned at a considerable distance and influenced by light. The \textcolor{myyellow}{Yellow Boxes} in the middle column denote failed predictions, including false alarms and missed detections. The \textcolor{green}{Green Boxes} and \textcolor{myblue}{Blue Boxes} in the right column indicate GT for known and unknown classes, respectively.}
    \label{vis3}
\end{figure}

\textbf{Failure Analysis.} 
Despite bringing about benefits, event cameras are not devoid of imperfections. Fig. \ref{vis3} reveals some failure cases attributed to the inherent characteristics of event cameras. For instance, the false alarm in Fig. \ref{vis3:event} results from the absence of color information, causing them to identify certain contour-similar backgrounds as objects. In addition, the sparse nature of events further aggravates the difficulty in detecting small objects with inherently limited appearance information, as illustrated by the missed detection in Fig. \ref{vis3:event}. Moreover, when dealing with distant and light-affected small targets as depicted in Fig. \ref{vis3:ori}, both event-based prediction and GT fail. 

\section{Conclusion}
\label{sec:conclusion}

In this paper, we have made the first exploration of the event-based vision in an open-set scenario, and developed the DEOE approach, which provides robustness against fast-moving objects and illumination variance. To identify objects with unknown categories, we propose to leverage the spatio-temporal consistency of the event stream's predictions and disentangle the unknown object discovery task with the foreground-background classification task. Built upon the state-of-the-art high-speed backbone RVT, our DEOE outperforms three strong baselines across multiple evaluation settings, showing better generalization ability to localize objects with novel categories. We anticipate this work will pave the way for further research, encompassing the development of open-set detection algorithm and the establishment of large-scale event camera open-set detection datasets covering a broader range of categories.

\ifCLASSOPTIONcaptionsoff
  \newpage
\fi

\bibliographystyle{IEEEtran}
\bibliography{papername,main}

\end{document}